%% file: main.tex
\documentclass{article}

\input{preamble}

\begin{document}

\title{Multimodal Side-Tuning for Document Classification}

\author{Stefano Pio Zingaro \and Giuseppe Lisanti \and Maurizio Gabbrielli}

\date{}

\maketitle
\begin{abstract}
In this paper, we propose to exploit the side-tuning framework for multimodal document classification.
Side-tuning is a methodology for network adaptation recently introduced to solve some of the problems related to previous approaches. 
Thanks to this technique it is actually possible to overcome model rigidity and catastrophic forgetting of transfer learning by fine-tuning.
The proposed solution uses off-the-shelf deep learning architectures leveraging the side-tuning framework to combine a base model with a tandem of two side networks.
We show that side-tuning can be successfully employed also when different data sources are considered, e.g.\ text and images in document classification.
The experimental results show that this approach pushes further the limit for document classification accuracy with respect to the state of the art.
\end{abstract}
\section{Introduction}\label{sec:introduction}
\input{introduction}
\section{Related Work}\label{sec:related}
\input{related}
\section{Methodology}\label{sec:method}
\input{method}
\section{Experimental Results}\label{sec:experiment}
\input{experiment}
\section{Conclusion}\label{sec:conclusion}
\input{conclusion}
\section*{Acknowledgments}
The Titan Xp GPU used for this research was donated by the NVIDIA Corporation.
\bibliographystyle{IEEEtran}
\bibliography{IEEEabrv,main}
\end{document}

%% file: preamble.tex
\usepackage[T1]{fontenc}
\usepackage[utf8]{inputenc}
\usepackage{mathptmx}
\usepackage{amsthm}
\usepackage{enumitem} 
\makeatletter
\let\NAT@parse\undefined{}
\makeatother
\usepackage[
    ]{hyperref}
\usepackage{xspace}
\usepackage{cite}
\usepackage{graphicx}
\graphicspath{{assets/}}
\DeclareGraphicsExtensions{.pdf,.jpeg,.png}
\usepackage{amsmath}
\usepackage{array}
\usepackage[caption=false,font=footnotesize]{subfig}
\usepackage{xcolor}
\usepackage{booktabs}
\usepackage[subtle]{savetrees}

%% file: introduction.tex
%
Notwithstanding the many technological advances in computer vision and artificial intelligence, which are contributing to the  ``digital transformation'' of many companies and industrial processes, there still exist a surprising number of tasks which are almost completely carried out by humans. 
In particular, many tasks in different industries, from administrative procedures to archival of old manuscripts, involve the human elaboration of a huge number of paper documents, with consequent high costs for the companies and, ultimately, for their clients.
There are two main reasons for this situation: one is deeply connected to the internal rules and processes of some companies, banks in particular, which have an important number of legacy procedures and have big inertia for innovation.
The second reason, that we consider in this paper, is the lack of completely satisfactory (automatic) tools for document classification, especially when documents contain different source of information such as text, images, and handwritten parts. 
While some paper documents could be replaced by electronic means, one cannot eliminate paper documentation, hence efficient and trustworthy tools for document classification are essential.

As we discuss in the next section, document classification has been widely investigated and methods can be roughly divided into three categories: those that are based on the textual content of the document, often obtained from \textit{Optical Character Recognition} (OCR), those based on the visual structure of the image, and multimodal methods that use both text and image.
The latter family of solutions~\cite{6831020, bagdanovIJDAR2014, 10.1145/2960811.2960814, wiedemann2017page, enginmultimodal, 8977998, 8978000, audebert} have provided significant advances, yet dealing with both textual and visual content in full generality remains an open problem~\cite{audebert}.

In this paper, we tackle the challenge by exploiting \textit{side-tuning}~\cite{zhang2019side} --- a recent methodology for network adaptation --- in multimodal document classification.
In general, network adaptation is a common technique that allows updating the weights of a pre-trained model on a different task. 
This technique is opposed to training from scratch and allows, among other benefits, a faster convergence.
However, existing adaptation solutions may suffer from catastrophic forgetting that is, the tendency of a network to abruptly lose previous knowledge when learning new information.
Side-tuning~\cite{zhang2019side} addresses the problem of adaptation by using a second network  whose weights are never updated, so as to preserve the classification capability of the original task.
The output of the base network and the side network are then merged into a specific layer.
The fusion takes place using an appropriate sum operation of the single outputs\footnote{Several notions of summation can be used, details can be found in~\cite{zhang2019side}.}.
Similarly to other additive learning approaches, side-tuning does not change the base model, rather it adapts it to a new target task by adding new parameters.
However, differently from other approaches, side-tuning does not impose any constraints on the structure of the side network, whose complexity can be scaled to the difficulty of the problem at hand, thus allowing also tiny networks when the base requires minor updates. 
This provides an extreme flexibility of the model and it is one of the reasons for its good results.

Our research idea is to exploit side-tuning also in the field of multimodal document classification, based on the intuition that this enhanced flexibility could allow one to precisely tune the model on different sources (i.e., textual and visual), while avoiding catastrophic forgetting and model rigidity.
We implement our idea by proposing a new method for multimodal learning with a deep neural network model, more precisely we present a side-tuned architecture that uses off-the-shelf networks and consists of one base model with a tandem of two side networks. 
Our experimental results show that this architecture is effective in common document classification scenarios and pushes further the limit for document classification accuracy.

The remaining of the paper is organized as follows, Section~\ref{sec:related} reviews related work and discuss the contributions of our solution.
Section~\ref{sec:method} explains the methodology and provides details concerning the model implementation. In Section~\ref{sec:experiment}, we provide the results of the experimental procedure used to assess the model validity and compare those results with previous works discussing the implementation choices.  
Finally, Section~\ref{sec:conclusion} summarizes the contributions and addresses some future directions for the presented work.

%% file: related.tex
%
Document classification has been widely investigated and several solutions have been proposed over the years. 
These solutions can be categorized considering whether they analyze the textual content of a document, its visual structure or both.
A complete analysis on text classification methods before the rise of deep learning solutions can be found in~\cite{aggarwal2012survey}.
Recently, Kim~\cite{kim2014convolutional} proposed to use Convolutional Neural Networks (CNNs) on top of a pre-trained embedding to perform sentence classification providing an effective and portable solution that has been widely used in many subsequent work~\cite{kim2014convolutional, joulin2016bag, yang2016hierarchical, bojanowski2017enriching, 10.5555/3304222.3304391}.
In~\cite{qiu2020pre} the authors give a thorough review on pre-trained models for natural language processing.

In the past, classification of a document based on its visual content has always been addressed with the design of hand-crafted features. 
These features were used to extract meaningful information about the image content or the document structure and then used as input to classic machine learning techniques for classification. 
A thorough analysis of these solutions can be found in the survey by Chen and Blostein~\cite{visual-survey}.
However, the recent advances in document image classification have been mostly led by solutions exploiting CNNs~\cite{6977258, 7333910, 7333933, csurka2016right, 8270002, 8270080, 8270148, 8545630}. 
Kang et al~\cite{6977258} proposed the first solution based on CNNs for document images classification. They designed a shallow architecture composed by two convolutional layers, max pooling and two fully connected layers, with ReLU activations and dropout regularization. The network was trained from scratch and the final results showed the superior performance of CNNs compared to classic solutions~\cite{visual-survey}.
The solutions proposed in~\cite{7333910, 7333933} demonstrated that it is possible to further improve this performance by exploiting transfer learning. In both articles, the authors successfully fine-tuned a state-of-the-art architecture, such as AlexNet~\cite{alexnet} (previously trained on ImageNet~\cite{imagenet_cvpr09}), to recognize the document type. 
Successively, the authors of~\cite{8270002} performed a thorough analysis on how different image pre-processing steps and architecture hyper-parameters may affect the final classification performance. They performed several tests, training each networks from scratch, and obtained  results comparable to the previous solutions.
In~\cite{8270080} several state-of-the-art very deep architectures, such as VGG16, GoogLeNet and ResNet-50 have been trained and/or fine-tuned for recognizing document images, achieving a huge boost in performance. 
Differently from the previous approaches, the solution in~\cite{8270148} exploited pre-trained CNNs just to extract the features from document images and then used extreme learning machines (ELMs) for classification.
The solution in~\cite{8545630} performed two steps of fine-tuning. In particular, given a pre-trained VGG16 architecture, a first fine-tuning is performed exploiting the whole visual content of document images. Then a second transfer learning is performed on specific image regions. 
Finally, the results is obtained as the combination of the predictions from all these neural network models.

Several papers proposed to combine both textual and visual features for documents classification~\cite{6831020, bagdanovIJDAR2014, 10.1145/2960811.2960814, wiedemann2017page, enginmultimodal, 8977998, 8978000, audebert}. 
The method in~\cite{6831020} combined bag-of-words and bag-of-visual-words representations exploiting SVM and a late fusion scheme. 
Similarly, in~\cite{bagdanovIJDAR2014} the authors used a bag-of-words representation with latent semantic analysis for the text and the visual descriptor from~\cite{711385} for images. Different classifiers with both early and late fusion schemes have been used to combine the text and visual features in order to correctly classify a page stream. 
In~\cite{10.1145/2960811.2960814} the document was first processed by an OCR\@. Successively, the extracted words were highlighted in the original document image through colored bounding boxes, following a ranking algorithm. These newly generated images were used to train a CNN for classification. 
The solution proposed in~\cite{8977998} tested two different fusion schemes, in particular, a spatial fusion and a features fusion scheme. 
In the spatial fusion, text and images are concatenated and given as input to a VGG16 network for training. Whereas, in the features fusion, the image feature obtained from a VGG16 network and text feature obtained through a text ensemble network are stacked and fed to a fully connected layer for classification.
Similarly, the authors of~\cite{8978000, audebert} proposed two solutions, which differ mainly in the embedding used for text and the CNN architectures used for images, InceptionV3 and MobileNetV2 respectively. 
As in~\cite{8977998}, the features extracted from text and image networks are concatenated and fed to a fully connected layer for classification. 
Both architectures have been trained end-to-end and have achieved state-of-the-art performance.

\subsection*{Contributions}
Differently from the previous approaches, we combine incremental learning and multimodal features training to jointly learn from both representations, visual and textual.
The resulting model presents great flexibility and keeps high performance when used on both small and large datasets.
To the best of our knowledge, our approach is the first that successfully attempts to apply side-tuning by using different sources of input during training.
We thoroughly evaluate our approach on two publicly available datasets~\cite{KUMAR2014119,7333910} and two different deep learning architectures in order to assess the validity of the proposed model.
The final model performance is competitive with state-of-the-art solutions on both datasets.

%% file: method.tex
%
In this section, we provide the details of a multimodal document classification model that takes advantage of side-tuning to properly combine visual and textual features. 
In the side-tuning framework, architectural elements are combined to produce a new representation of the target~\cite{zhang2019side}.
A side-tuning architecture generally presents a base model with fixed weights and a side model whose weights are unlocked to allow updating.
In principle, different architectures can be selected for the base and side models to allow modularity of the components. 
For example, the authors of~\cite{zhang2019side} use the concept of knowledge distillation for neural networks~\cite{Hinton2015} to properly initialize the weights of the side component architecture.

In the implementation discussed in this paper, the base model consists of a  Convolutional Neural Network (CNN) for image classification, pre-trained on the ImageNet dataset.
The side component presents two different networks: the first one is identical to the base model but with unlocked weights to allow update during training, while the second network is a CNN for text classification.
In defining the final model, we can rely on two strategies.
The first involves the distillation of a network, while the second uses networks as they are. 
We choose the latter and select small network architectures so that we do not have to compress the model for image classification.
\begin{figure*}[!t]
    \centering
    \includegraphics[width=\linewidth]{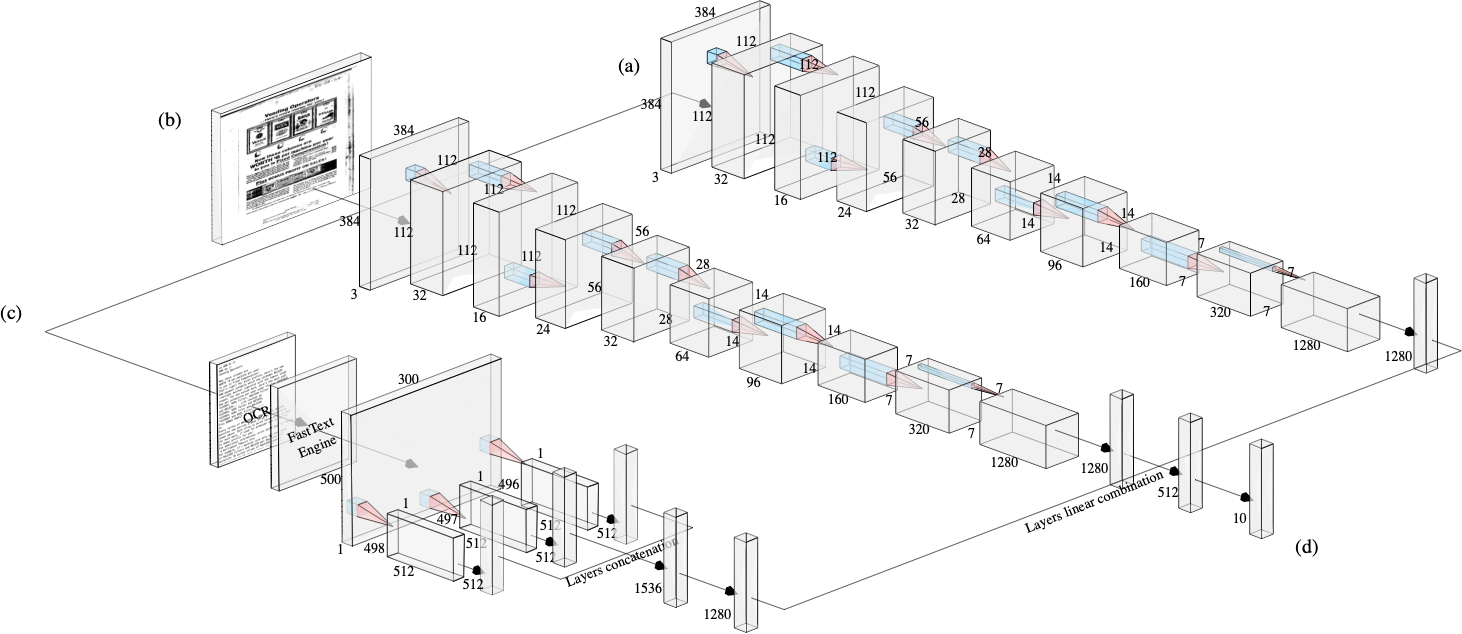}
    \caption{Multimodal side-tuning classifier for hybrid text and image classification. Base model (a) and side model (b) reflects the same MobileNetV2 architecture, while (c) is a CNN inspired from sequence text classification task. The final merge architecture combines the output of the three networks into one new encoding as shown in (d).}\label{fig:method_model_overview}
\end{figure*}

In the remaining of the section we provide the networks details of the baseline models for both images and text and then we describe the multimodal combination process.

\subsection{Model for visual features}\label{subsec:method_image}
Deep Convolutional Neural Networks (DCNN) have proven to be effective when pre-trained on large dataset and successively fine-tuned for a different task using a smaller set of data~\cite{Yosinski2014}.
We considered two DCNNs pre-trained on the ImageNet dataset as the reference architectures for the document image classification.
As a first attempt in the definition of the model, we choose the MobileNetV2~\cite{Sandler2018} neural network.
The MobileNet networks family has been originally designed to exploit Deep Learning on resource-constrained devices.
Its relatively simple architecture presents a smaller number of trainable parameters (about $3.5$M) and yet it achieves competitive classification performance with respect to more complex and resource-greedy models~\cite{alexnet,Szegedy2015,Simonyan2015,Szegedy2016}.
The learning process of MobileNetV2 is based on the principle of learning residuals and uses the combination of expansion levels and bottleneck blocks to effectively encode the image features.
Despite of the specific reasons to select the MobileNetV2 architecture, we have also considered the ResNet50 model. 
In principle, we could have employed any other popular DCNN, e.g.\ VGG16, InceptionV3, to accomplish the image classification task.

We pre-process the network input by resizing the image to $384 \times 384$ and by replicating the grayscale to respect the original network input that is, three channels RGB images.
As a consequence of the adoption of the ImageNet pre-trained model, we centre the input by applying standardization using mean and variance values from the training dataset.

\subsection{Model for textual features}\label{subsec:method_text}
The classification of documents from scans presenting hybrid text/image content involves the creation of a corpus including the textual version of each input image.
The corpus should then be coded in an appropriate format using, for instance, an approach similar to the \textit{word2vec} model~\cite{Mikolov2013}.
It is appropriate to carefully select the specific model to be used for words vectorization since different implementation strategies could affect the quality of learning. 
Such choices comprise the measure for calculating the similarity distance between the words or the method for the vectors initialization.
The analysis of corpora generation strategies lays beyond the scope of this work, nevertheless, previous work addressed that this procedure is key to obtain good results~\cite{audebert}.
Furthermore, since this problem has already been addressed in the reference literature, text versions of the datasets considered in this work already exist: \textit{QS-OCR-Small} for \textit{Tobacco3482} and \textit{QS-OCR-large} for \textit{RVL-CDIP}~\cite{audebert}, obtained using the \textit{Tesseract} OCR 4.0 engine, which is based on LSTM~\cite{Smith2007}.

In Natural Language Processing (NLP) practice, vector form encoding involves a tokenization procedure followed by the creation of a lookup table that associates a unique numeric identifier to each word in the resulting vocabulary.
This embedding procedure aims to represent a text with a real-valued vector of numbers that is used in an end-to-end training to learn similarities among different words. 
In our case, the tokenization is carried out separating words by white spaces without ignoring punctuation, nor removing digits or OCR-produced artifacts.
This way we aim at exploiting, on the one hand, the OCR ``noise'' as a regularization factor for the training procedure and, on the other hand, the consistency of the OCR to recognize similar patterns.

Similarly to the image case, text classification also benefits of weights initialization from a large corpora of pre-trained models.
In fact, the creation of the lookup table can be replaced with an already existing vocabulary, which contains information on words similarity previously computed with proper distance measurements, e.g. Levenshtein in the case of \textit{GloVe}~\cite{Pennington2014} and \textit{ELMo}~\cite{Peters2018}.
Therefore, we choose a pre-trained model that contains embeddings for each word of our corpus, we combine all the vectors representing the words of a single text document, and we 0-pad the encodings which contain less than $500$ words, as in~\cite{audebert}.
Considering the characteristics described, we select \textit{FastText}~\cite{joulin2016bag} among the models in the literature. 
FastText is pre-trained on the \textit{Common Crawl} dataset~\cite{bojanowski2017enriching} and generates embeddings of $k=300$ real values per word. 
Remarkably, it is able to encode every token in the datasets considered in this work.
Indeed, we believe that avoiding models with \textit{Out-Of-Vocabulary} (OOV) words is crucial to exploit the embeddings in the procedure.

We carry out the baseline training for the text classification model with a simple architecture (about $1.8$M parameters) inspired by a CNN for sentence classification~\cite{kim2014convolutional}.
The network consists of three convolutional layers of dimension $h \times k$, each starting from the same input and acting in parallel.
The convolutional layers use a window size of $h={3,4,5}$ words, no padding and a stride of $1$. 
Each layer has $512$ filters, uses \textit{ReLU} activation function and a resizing step with one-dimensional \textit{max-pooling}.
The resulting tensors are concatenated and fed to a classification layer with Softmax activation.
We also apply a dropout regularization with a fixed probability of $0.5$.
As in the case of the model for visual features, we could have chosen any other off-the-shelf architecture.

\subsection{Combined model}
To benefit from both representations we choose to combine image and text in a single, new, encoding.
In our setup, we use a network with locked weights and a side model, which is composed of the two architectures described in Subsections~\ref{subsec:method_image} and~\ref{subsec:method_text} without the final classification layer.
The base and side networks that take the image as input are pre-trained on ImageNet, while the weights of the side network for text classification are randomly initialized.
The combination of the three encodings can be addressed with different methods, of which we list the two most significant.
First, we can concatenate the outputs, delegating the task of selecting the most significant weights to the fusion network. 
Second, we can linearly combine the encodings so as to align the feature space and select the best coefficients.

The first concatenation method have been exploited in several works~\cite{audebert,Eitel2015,8978000}, all reporting an increase in accuracy performance with respect to the single baseline models.
The second method is less explored and advocates for a linear merging of the encodings.
Concretely, the combination of the base and side models in our architecture is performed as:
\begin{equation}
    \label{eq:method_merge}
    R(x) = \alpha_0 B(x) + \sum_{i=1}^{N}{\alpha_{i}S_{i}(x)}, 
\end{equation}
where $R$ is the new representation for the given task, $B$ and $S_i$ are respectively the base and sides model encodings, and $\alpha_i$ are coefficients of the equation, subject to the constraint
\begin{math}
    \sum_{i=0}^{N}{\alpha_{i}=1}
\end{math}\@.
In our case, where $N=2$, the overall combination assumes the form
\begin{math}
    \alpha_0 B(x) + \alpha_1 S_1(x) + \alpha_2 S_2(x)
\end{math}\@.
It is worth noting that some specific values for the alpha coefficients lead to well-known training procedures, that in our case corresponds to just the image feature extraction ($\alpha_0=1$, $\alpha_1=0$, and $\alpha_2=0$), to the fine-tuning of the image architecture ($\alpha_0=0$, $\alpha_1=1$, and $\alpha_2=0$), and finally to the training from scratch of the text network ($\alpha_0=0$, $\alpha_1=0$, and $\alpha_2=1$).
Setting properly these coefficients allows to easily switch between the different modalities with a gain in flexibility and the possibility to explore  their combination.

In order to perform the weighted sum of the network outputs each resulting vector must have the same dimension.
In case of different input sources, it may be necessary to use an adaptation layer to make the output shapes compatible.
In our case, we use such layer to adapt the text output with the image one.
Finally, the result of the linear combination is passed to a classification layer. 
In addition to the architecture just described, we have performed experiments by adding a fully connected layer after the fusion and before the classification, to analyze the behavior of the model as the parameters increase.
An overview of the architecture is shown in Figure~\ref{fig:method_model_overview}.

%% file: experiment.tex

\begin{figure*}[ht]%
\centering
\subfloat[][]{
    \includegraphics[width=0.46\textwidth]{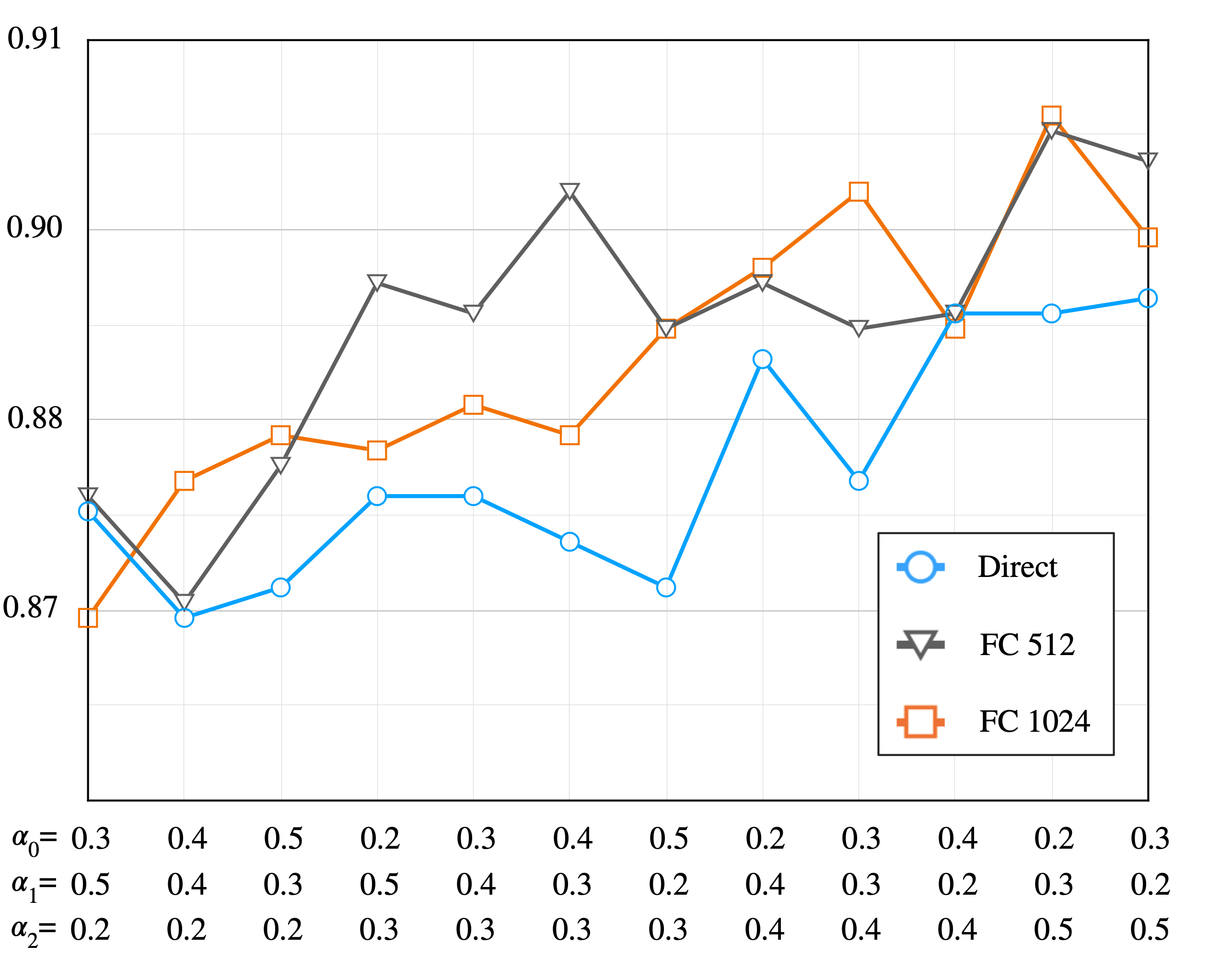}
}%
\qquad
\subfloat[][]{
    \includegraphics[width=0.46\textwidth]{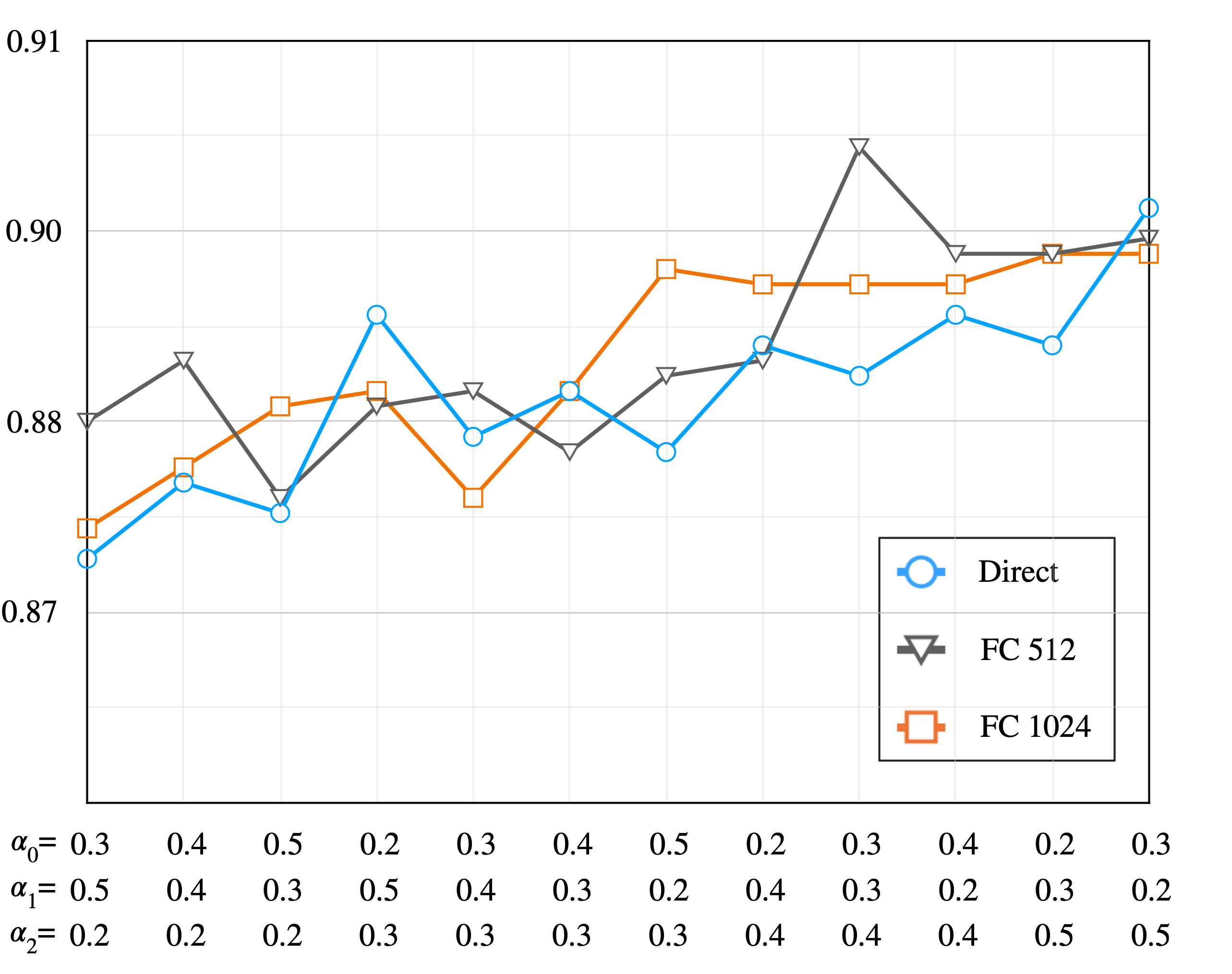}
}
\caption{Plot of the accuracy on the Tobacco3482 dataset for twelve combinations of the multimodal coefficients. Base and first side component use MobileNetV2 (a) and ResNet50 (b) architectures. Each trend corresponds to a different configuration of the side-tuned network.}\label{fig:exp_coefficients}%
\end{figure*}
We performed an analysis of the multimodal side-tuning architecture to assess the quality of our methodology and to better understand how it contributed to the classification accuracy.
In the following, we first introduce the datasets used in our experiments, then we detail the training procedure, and finally we provide a comparison of the performance with respect to the state-of-the-art.
We also give a brief analysis of the inference process running time.

\subsection{Datasets}
The Tobacco3482 dataset~\cite{KUMAR2014119} comprises $3482$ greyscale scans of documents divided unevenly in $10$ categories, e.g.\ resume, email, letter, memo.
The documents distribution among the classes spans from $120$ for the resume category to $620$ for memo.
It is a small subset of the Truth Tobacco Industry Documents and collects many hybrid content documents.
The textual version of this dataset, namely \textit{QS-OCR-Small}~\cite{audebert} reflects the same structure of the original image dataset. 
In our setting, we random sampled three subsets to be used for train, validation and test, fixing their cardinality to $800$, $200$, and $2482$ respectively, as in~\cite{7333910,audebert}.

The Ryerson Vision Lab Complex Document Information Processing (RVL-CDIP) dataset~\cite{7333910} contains $399828$ images divided into $16$ categories from the Truth Tobacco Industry Documents, e.g.\ scientific publication, scientific report, handwritten. 
Its textual counterpart, \textit{QS-OCR-Large}, was developed in the same work that released \textit{QS-OCR-Small}~\cite{audebert}.
Differently from Tobacco3482, RVL-CDIP comes with pre-built subsets for train, validation and test that have respectively dimension of $319837$, $39995$, and $39996$.

\begin{table}[t]
    \caption{Baseline models and multimodal overall accuracy for Tobacco3482 using MobileNetV2 (visual features) and 1D CNN (textual features) architectures. Best result in bold.}\label{tab:exp_baseline}
    \centering
    \begin{tabular}{llc}
        \toprule
        \bfseries Model & \bfseries \#Params & \bfseries OA \\ 
        \midrule
        \textit{Text} & $\approx 1.8$M & $67.8\%$ \\
        \textit{Image} ({\scriptsize fine-tuning}) & $\approx 3.5$M & $84.0\%$ \\    
        \textit{Image} ({\scriptsize side-tuning}) & $\approx 7$M & $88.0\%$ \\
        \midrule
        \textit{Multimodal} ({\scriptsize side-tuning}) & $\approx 12$M & $\mathbf{90.5\%}$ \\
        \bottomrule
    \end{tabular}
\end{table}

\subsection{Training details}\label{subsec:exp_details}
All the models are implemented using the \textit{PyTorch} framework, version $1.4.0$, and trained using an \textit{NVIDIA Titan XP} GPU\@. 
The hyper-parameters are selected from the experiments performed on the Tobacco3482 dataset.

We set the maximum number of epochs to $100$ and the batch size to $16$ documents for Tobacco3482 experiments while we chose to train, validate, and test with batches of $40$ for RVL-CDIP for $10$ epochs. 
We used the cross-entropy loss function in all the experiments.

We performed all tests using the \textit{Stochastic Gradient Descent} (SGD) optimizer with a momentum of $0.9$ and an initial learning rate of $0.1$, subject to a scheduled update at each iteration that follows the scheme proposed in~\cite{8270080}:
\begin{equation}
    \mathsf{LearningRate}_i = 0.1 * \sqrt{\frac{\mathsf{Epoch}_i}{\mathsf{MaxEpoch}}} 
    \label{eq:scheduler}
\end{equation}

\subsection{Ablation Study}\label{subsec:exp_ablation}
Table~\ref{tab:exp_baseline} reports the \textit{Overall Accuracy} (OA) obtained on the Tobacco3482 dataset.
As shown in the table, the multimodal network outperforms the other combinations, proving that we are able to combine efficiently the features space and benefit from side-tuning.
The side-tuning architecture used in Table~\ref{tab:exp_baseline} uses the MobileNetV2 as base model and side component for visual features. 
The model is pre-trained on ImageNet, coefficients are $\alpha_0=0.5$ for image-only side-tuning, while $\alpha_0=0.2$, $\alpha_1=0.3$, and $\alpha_2=0.5$ for the multimodal version.

The second analysis explores the behavior of multimodal side-tuning with respect to different coefficients for the linear combination.
Indeed, each $\alpha_i$ plays a central role in the balancing of the learning process for side-tuning.
To assess their impact in our setting, we train several models following twelve different alpha configurations.
We select values ranging from $0.1$ to $0.5$ to always be able to exploit each component of the framework without excessively lowering the weights of the other networks.
We also consider two architectures for the image input (MobileNetV2 and ResNet50) and for both we tested two different network configurations.
The first inputs directly the combination of the base model and side models to the classification layer, while the second considers an additional FC layer before the classification one.
We test two different dimensions for the latter.
In Figure~\ref{fig:exp_coefficients}, we analyze the behaviors of this set of experiments.

The coefficients are ordered so that the linear combination in the merging layer gives incrementally more importance to the model component exploiting textual features ($\alpha_2$).
The accuracy increases with the progressive shift from the model that favors visual features, with $\alpha_0$ or $\alpha_1$ greater than $\alpha_2$, to a more text-centered classifier.
Small changes in the coefficients affect the training for all the architectures.
Nevertheless, those models with the additional fully connected layers, both in MobileNetV2 and in ResNet50 show the best trends. 
In particular, the best accuracy is reached by model with a dense layer of dimension 1024 for MobileNetV2 with the configuration $\alpha_0=0.2$,  $\alpha_1=0.3$, and  $\alpha_2=0.5$ and the one with a dense layer of dimension 512 for ResNet50 with the configuration $\alpha_0=0.3$,  $\alpha_1=0.3$, and  $\alpha_2=0.4$.
In Table~\ref{tab:exp_architectures} we present the best results for MobileNetV2 and ResNet50 architectures on the Tobacco3482 dataset.
First, we tested the two architecture in the side-tuning framework using one side component and the same alpha configuration ($\alpha_0=0.5$ and  $\alpha_1=0.5$).
Next, we take advantage of the second side component --- the text classifier --- to perform the multimodal side tuning.
Although very similar, both the experiments present better results when the MobileNetV2 architecture is selected.

In Table~\ref{tab:exp_overall_acc_rvl} we report the experiments for RVL-CDIP dataset, presenting a different trend with respect to Tobacco3482 experiments, in fact, ResNet50 has the best accuracy.
This is due to the fact that an architecture with a larger number of parameters (ResNet50) can benefit from a bigger dataset (RVL-CDIP) while suffering from small inter-class variability.
\begin{table}[t]
    \caption{Overall accuracy on the Tobacco3482 image dataset using two different off-the-shelf architectures for both base and side model in the side tuning framework. Best result in bold.}\label{tab:exp_architectures}
    \centering
    \begin{tabular}{llcc}
        \toprule
        \bfseries Model (base architecture) & \bfseries \#Params & \bfseries OA \\ 
        \midrule
        \textit{Image (ResNet50)} & $\approx 51$M & $87.2\%$ \\
        \textit{Image (MobileNetV2)} & $\approx 7$M & $88.0\%$ \\    
        \midrule
        \textit{Multimodal (ResNet50)} & $\approx 57$M & $90.3\%$ \\
        \textit{Multimodal (MobileNetV2)} & $\approx 12$M & $\mathbf{90.5\%}$ \\    
        \bottomrule
    \end{tabular}
\end{table}
\begin{table}[t]
    \caption{Overall accuracy on the RVL-CDIP dataset compared with the results from previous works. Modalities of the data source are  image (\texttt{I}), text (\texttt{T}), or both (\texttt{I+T}). The selected alpha configuration  for the multimodal side-tuning is $a_0=0.3$, $a_1=0.2$, and $a_2=0.5$ for MobileNetV2 and $a_0=0.3$, $a_1=0.3$, and $a_2=0.4$ for ResNet50. Best result in bold.}\label{tab:exp_overall_acc_rvl}
    \centering
    \begin{tabular}{lccl}
        \toprule
        \bfseries Model & \bfseries \#Params & \bfseries Modality & \bfseries OA \\
        \midrule
        \textit{CNNs}~\cite{7333910} & $\approx 62$M & \texttt{I} & $89.8\%$ \\
        \textit{Audebert}~\cite{audebert} & $\approx 8$M & \texttt{I+T} & $90.6\%$ \\
        \textit{AlexNet + SPP}~\cite{8270002} & $\approx 62$M & \texttt{I} &$90.94\%$ \\
        \textit{VGG16}~\cite{8270080} & $\approx 138$M & \texttt{I} & $90.97\%$ \\
        \textit{VGG16 + ULMFit}~\cite{8977998} & $\approx 162$M & \texttt{I+T} & $\mathbf{93.6\%}$ \\
        \midrule
        \textit{Text} & $\approx 1.8$M & \texttt{T} & $80.5\%$ \\
        \textit{Multimodal (MobileNetV2)} & $\approx 12$M & \texttt{I+T} & $92.2\%$ \\    
        \textit{Multimodal (ResNet50)} & $\approx 57$M & \texttt{I+T} & $92.7\%$ \\
       \bottomrule
    \end{tabular}
\end{table}
\begin{table*}[ht]
    \caption{Overall and per-class accuracy on the Tobacco3482 dataset compared with the results from~\cite{audebert}. The selected alpha configuration  for the multimodal side-tuning is $a_0=0.3$, $a_1=0.2$, and $a_2=0.5$ for MobileNetV2 and $a_0=0.3$, $a_1=0.3$, and $a_2=0.4$ for ResNet50. Best results in bold.}\label{tab:exp_tobacco_class}
    \centering
    \begin{tabular}{lccccccccccc}
        \toprule
        \bfseries Model & \bfseries OA & \bfseries Adve & \bfseries Email & \bfseries Form & \bfseries Letter & \bfseries Memo & \bfseries News & \bfseries Note & \bfseries Report & \bfseries Resume & \bfseries Scientific\\
        \midrule
        \textit{Audebert} & $87.8\%$ & $93.0\%$ & $98.0\%$ & $88.0\%$ & $86.0\%$ & $90.0\%$ & $90.0\%$ & $85.0\%$ & $71.0\%$ & $86.0\%$ & $68.0\%$ \\
        \midrule
        Text & $67.8\%$ & $93.3\%$ & $29.5\%$ & $77.0\%$ & $58.8\%$ & $49.7\%$ & $63.6\%$ & $68.7\%$ & $52.0\%$ & $60.7\%$ & $\mathbf{79.9\%}$ \\
        \textit{Multimodal (ResNet50)} & $90.3\%$ & $\mathbf{96.1\%}$ & $98.3\%$ & $\mathbf{90.8\%}$ & $91.7\%$ & $\mathbf{93.5\%}$ & $\mathbf{95.5\%}$ & $87.6\%$ & $\mathbf{76.7\%}$ & $89.4\%$ & $68.0\%$ \\
        \textit{Multimodal (MobileNetV2)} & $\mathbf{90.5\%}$ & $94.8\%$ & $\mathbf{99.1\%}$ & $88.7\%$ & $\mathbf{93.2\%}$ & $93.0\%$ & $\mathbf{95.5\%}$ & $\mathbf{89.7\%}$ & $76.2\%$ & $\mathbf{95.3\%}$ & $67.4\%$ \\
        \bottomrule
    \end{tabular}
\end{table*}

\subsection{Comparison with the state-of-the-art}
We proved the effectiveness of multimodal side-tuning compared to the fine-tuning on images and training from scratch on textual features.

In Table~\ref{tab:exp_overall_acc_rvl}, we compare five different state-of-the-art solutions with the proposed multimodal approach in terms of  overall accuracy on the RVL-CDIP dataset.
All the experiments have been carried out considering only the best configurations of alpha for both MobileNetV2 and ResNet50.

The works considered are the CNN implementation of~\cite{7333910}, the multimodal solution from~\cite{audebert}, the VGG16 network in~\cite{8270080}, the AlexNet implementation of~\cite{8270002}, and the VGG16 $+$ UMLFit of~\cite{8977998}. 
As it is possible to observe, performance on the RVL-CDIP dataset highlights that the proposed solution slightly improves the classification performance with respect to the methods proposed in~\cite{7333910,audebert,8270080,8270002} but obtains slighter lower results compare to~\cite{8977998}.
This is related to the difference in the networks complexity between our solution and the method from~\cite{8977998}, which has $+150$M parameters.

Finally, among these solutions, we select the one from~\cite{audebert}, the most similar approach to what we propose, and compare the per-class accuracy on the Tobacco3482 dataset.
In fact, the authors in~\cite{audebert} strived to use lightweight architecture as in our case but concatenated the output of the networks used for the images and the text before the classification. 
This also give us the chance to provide insights on the performance for the classes of interest in the Tobacco3482 dataset.
Table~\ref{tab:exp_tobacco_class} shows that the gain of our model is consistent over all the classes except for the scientific class. 
When compared with the solution proposed in~\cite{audebert}, the side-tuning model improves the overall accuracy of $2.7\%$.

\subsection{Processing time}\label{subsec:exp_time}
We now provide a discussion about execution time of our algorithm to analyze the performance of the document classification system.
Although some document analysis could be conducted offline, critical applications require low latency in order to be performed as close to real time as possible.
We then averaged the timings of the multimodal MobileNetV2 version over five classification runs.
The full inference process of our model on a single document is carried on a Intel Xeon Silver 4208 CPU takes $\approx 1595$ms. 
Of those, $\approx 910$ms ($57\%$) are spent for Tesseract OCR image processing and text extraction\footnote{Average timing for Tesseract has been computed using four threads as in~\cite{audebert}.}, $\approx 166$ms ($10.4\%$) for evaluation of the base model, $\approx 202$ms ($12.7\%$) for the side model exploiting image features, and $\approx 16$ms ($1\%$) are spent in the inference of the side component fed with textual features.
The timings for the image ($\approx 119$ms) and text load ($\approx 182$ms) from disk occupy the remaining time ($18.9\%$).
On NVIDIA Titan Xp GPU, the side-tuning model runs in $\approx 1224$ms --- whit Tesseract OCR occupying $74.3\%$ of the time. The base model is evaluated in $\approx 8$ms, $\approx 15$ms for the image side model, and $\approx 5$ms for the text model.

Compared to models with more complex architectures, the proposed system is able to be used in real-time applications with latencies around the second. If the selected components were to be replaced with heavier models, this would lead to an inevitable performance impoverishment.

%% file: conclusion.tex
In this work we presented a multimodal approach for document classification  that takes into consideration both visual ant textual features classify a document.
We leverage the work done in the last state-of-art solutions for incremental learning and take advantage of the side-tuning framework to develop an hybrid architecture that performs on par with existing more complex solutions and outperforms similar lightweight approaches.
To further improve the performance, we aim at automatically tuning the coefficients used in the linear combination of both the base and sides models. 
We also want to investigate the possibility of exploiting an ensemble of text embeddings and combine them using the side-tuning framework.

%% file: main.bbl
\begin{thebibliography}{10}
\providecommand{\url}[1]{#1}
\csname url@samestyle\endcsname
\providecommand{\newblock}{\relax}
\providecommand{\bibinfo}[2]{#2}
\providecommand{\BIBentrySTDinterwordspacing}{\spaceskip=0pt\relax}
\providecommand{\BIBentryALTinterwordstretchfactor}{4}
\providecommand{\BIBentryALTinterwordspacing}{\spaceskip=\fontdimen2\font plus
\BIBentryALTinterwordstretchfactor\fontdimen3\font minus
  \fontdimen4\font\relax}
\providecommand{\BIBforeignlanguage}[2]{{%
\expandafter\ifx\csname l@#1\endcsname\relax
\typeout{** WARNING: IEEEtran.bst: No hyphenation pattern has been}%
\typeout{** loaded for the language `#1'. Using the pattern for}%
\typeout{** the default language instead.}%
\else
\language=\csname l@#1\endcsname
\fi
#2}}
\providecommand{\BIBdecl}{\relax}
\BIBdecl

\bibitem{6831020}
O.~{Augereau}, N.~{Journet}, A.~{Vialard}, and J.~{Domenger}, ``Improving
  classification of an industrial document image database by combining visual
  and textual features,'' in \emph{International Workshop on Document Analysis
  Systems}, 2014, pp. 314--318.

\bibitem{bagdanovIJDAR2014}
M.~Rusi\~{n}ol, V.~Frinken, D.~Karatzas, A.~D. Bagdanov, and J.~Llad\'{o}s,
  ``Multimodal page classification in administrative document image streams,''
  \emph{Int. J. Doc. Anal. Recognit.}, vol.~17, no.~4, pp. 331--341, Dec. 2014.

\bibitem{10.1145/2960811.2960814}
L.~Noce, I.~Gallo, A.~Zamberletti, and A.~Calefati, ``Embedded textual content
  for document image classification with convolutional neural networks,'' in
  \emph{ACM Symposium on Document Engineering}, 2016, pp. 165--173.

\bibitem{wiedemann2017page}
G.~Wiedemann and G.~Heyer, ``Page stream segmentation with convolutional neural
  nets combining textual and visual features,'' \emph{arXiv preprint
  arXiv:1710.03006}, 2017.

\bibitem{enginmultimodal}
D.~Engin, E.~Emekligil, M.~Y. Akp{\i}nar, B.~Oral, and S.~Arslan, ``Multimodal
  deep neural networks for banking document classification,'' in
  \emph{International Conference on Advances in Information Mining and
  Management}, 2019, pp. 21--25.

\bibitem{8977998}
R.~{Jain} and C.~{Wigington}, ``Multimodal document image classification,'' in
  \emph{International Conference on Document Analysis and Recognition (ICDAR)},
  2019, pp. 71--77.

\bibitem{8978000}
M.~N. {Asim}, M.~U.~G. {Khan}, M.~I. {Malik}, K.~{Razzaque}, A.~{Dengel}, and
  S.~{Ahmed}, ``Two stream deep network for document image classification,'' in
  \emph{International Conference on Document Analysis and Recognition (ICDAR)},
  2019, pp. 1410--1416.

\bibitem{audebert}
N.~Audebert, C.~Herold, K.~Slimani, and C.~Vidal, ``Multimodal deep networks
  for text and image-based document classification,'' in \emph{Machine Learning
  and Knowledge Discovery in Databases}, P.~Cellier and K.~Driessens,
  Eds.\hskip 1em plus 0.5em minus 0.4em\relax Springer International
  Publishing, 2020, pp. 427--443.

\bibitem{zhang2019side}
J.~O. Zhang, A.~Sax, A.~Zamir, L.~Guibas, and J.~Malik, ``Side-tuning: Network
  adaptation via additive side networks,'' \emph{arXiv preprint
  arXiv:1912.13503}, 2019.

\bibitem{aggarwal2012survey}
C.~C. Aggarwal and C.~Zhai, ``A survey of text classification algorithms,'' in
  \emph{Mining text data}.\hskip 1em plus 0.5em minus 0.4em\relax Springer,
  2012, pp. 163--222.

\bibitem{kim2014convolutional}
Y.~Kim, ``Convolutional neural networks for sentence classification,''
  \emph{arXiv preprint arXiv:1408.5882}, 2014.

\bibitem{joulin2016bag}
A.~Joulin, E.~Grave, P.~Bojanowski, and T.~Mikolov, ``Bag of tricks for
  efficient text classification,'' \emph{arXiv preprint arXiv:1607.01759},
  2016.

\bibitem{yang2016hierarchical}
Z.~Yang, D.~Yang, C.~Dyer, X.~He, A.~Smola, and E.~Hovy, ``Hierarchical
  attention networks for document classification,'' in \emph{Conference of the
  North American chapter of the association for computational linguistics:
  human language technologies}, 2016, pp. 1480--1489.

\bibitem{bojanowski2017enriching}
P.~Bojanowski, E.~Grave, A.~Joulin, and T.~Mikolov, ``Enriching word vectors
  with subword information,'' \emph{Transactions of the Association for
  Computational Linguistics}, vol.~5, pp. 135--146, 2017.

\bibitem{10.5555/3304222.3304391}
S.~Wang, M.~Huang, and Z.~Deng, ``Densely connected cnn with multi-scale
  feature attention for text classification,'' in \emph{International Joint
  Conference on Artificial Intelligence}, 2018, pp. 4468--4474.

\bibitem{qiu2020pre}
X.~Qiu, T.~Sun, Y.~Xu, Y.~Shao, N.~Dai, and X.~Huang, ``Pre-trained models for
  natural language processing: A survey,'' \emph{arXiv preprint
  arXiv:2003.08271}, 2020.

\bibitem{visual-survey}
N.~Chen and D.~Blostein, ``A survey of document image classification: Problem
  statement, classifier architecture and performance evaluation,'' \emph{Int.
  J. Doc. Anal. Recognit.}, vol.~10, no.~1, p. 1–16, May 2007.

\bibitem{6977258}
L.~{Kang}, J.~{Kumar}, P.~{Ye}, Y.~{Li}, and D.~{Doermann}, ``Convolutional
  neural networks for document image classification,'' in \emph{International
  Conference on Pattern Recognition}, 2014, pp. 3168--3172.

\bibitem{7333910}
A.~W. {Harley}, A.~{Ufkes}, and K.~G. {Derpanis}, ``Evaluation of deep
  convolutional nets for document image classification and retrieval,'' in
  \emph{International Conference on Document Analysis and Recognition (ICDAR)},
  2015, pp. 991--995.

\bibitem{7333933}
M.~Z. {Afzal}, S.~{Capobianco}, M.~I. {Malik}, S.~{Marinai}, T.~M. {Breuel},
  A.~{Dengel}, and M.~{Liwicki}, ``Deepdocclassifier: Document classification
  with deep convolutional neural network,'' in \emph{International Conference
  on Document Analysis and Recognition (ICDAR)}, 2015, pp. 1111--1115.

\bibitem{csurka2016right}
G.~Csurka, D.~Larlus, A.~Gordo, and J.~Almaz{\'a}n, ``What is the right way to
  represent document images?'' \emph{arXiv preprint arXiv:1603.01076}, 2016.

\bibitem{8270002}
C.~{Tensmeyer} and T.~{Martinez}, ``Analysis of convolutional neural networks
  for document image classification,'' in \emph{International Conference on
  Document Analysis and Recognition (ICDAR)}, 2017, pp. 388--393.

\bibitem{8270080}
M.~Z. {Afzal}, A.~{K\"{o}lsch}, S.~{Ahmed}, and M.~{Liwicki}, ``Cutting the
  error by half: Investigation of very deep cnn and advanced training
  strategies for document image classification,'' in \emph{International
  Conference on Document Analysis and Recognition (ICDAR)}, 2017, pp. 883--888.

\bibitem{8270148}
A.~{K\"{o}lsch}, M.~Z. {Afzal}, M.~{Ebbecke}, and M.~{Liwicki}, ``Real-time
  document image classification using deep cnn and extreme learning machines,''
  in \emph{International Conference on Document Analysis and Recognition
  (ICDAR)}, 2017, pp. 1318--1323.

\bibitem{8545630}
A.~{Das}, S.~{Roy}, U.~{Bhattacharya}, and S.~K. {Parui}, ``Document image
  classification with intra-domain transfer learning and stacked generalization
  of deep convolutional neural networks,'' in \emph{International Conference on
  Pattern Recognition (ICPR)}, 2018, pp. 3180--3185.

\bibitem{alexnet}
A.~Krizhevsky, I.~Sutskever, and G.~E. Hinton, ``Imagenet classification with
  deep convolutional neural networks,'' in \emph{Advances in Neural Information
  Processing Systems 25}, F.~Pereira, C.~J.~C. Burges, L.~Bottou, and K.~Q.
  Weinberger, Eds.\hskip 1em plus 0.5em minus 0.4em\relax Curran Associates,
  Inc., 2012, pp. 1097--1105.

\bibitem{imagenet_cvpr09}
J.~Deng, W.~Dong, R.~Socher, L.-J. Li, K.~Li, and L.~Fei-Fei, ``Imagenet: A
  large-scale hierarchical image database,'' in \emph{Conference on Computer
  Vision and Pattern Recognition (CVPR)}, 2009.

\bibitem{711385}
P.~Heroux, S.~Diana, A.~Ribert, and E.~Trupin, ``Classification method study
  for automatic form class identification,'' in \emph{Proceedings. Fourteenth
  International Conference on Pattern Recognition (Cat. No. 98EX170)},
  vol.~1.\hskip 1em plus 0.5em minus 0.4em\relax IEEE, 1998, pp. 926--928.

\bibitem{KUMAR2014119}
J.~Kumar, P.~Ye, and D.~Doermann, ``Structural similarity for document image
  classification and retrieval,'' \emph{Pattern Recognition Letters}, vol.~43,
  pp. 119--126, 2014.

\bibitem{Hinton2015}
G.~Hinton, O.~Vinyals, and J.~Dean, ``Distilling the knowledge in a neural
  network,'' \emph{arXiv preprint arXiv:1503.02531}, 2015.

\bibitem{Yosinski2014}
J.~Yosinski, J.~Clune, Y.~Bengio, and H.~Lipson, ``How transferable are
  features in deep neural networks?'' in \emph{Advances in Neural Information
  Processing Systems}, 2014.

\bibitem{Sandler2018}
M.~Sandler, A.~Howard, M.~Zhu, A.~Zhmoginov, and L.-C. Chen, ``Mobilenetv2:
  Inverted residuals and linear bottlenecks,'' in \emph{2018 IEEE/CVF
  Conference on Computer Vision and Pattern Recognition}.\hskip 1em plus 0.5em
  minus 0.4em\relax IEEE, Jun. 2018, pp. 4510--4520.

\bibitem{Szegedy2015}
C.~Szegedy, W.~Liu, Y.~Jia, P.~Sermanet, S.~Reed, D.~Anguelov, D.~Erhan,
  V.~Vanhoucke, and A.~Rabinovich, ``Going deeper with convolutions,''
  \emph{Proceedings of the IEEE Computer Society Conference on Computer Vision
  and Pattern Recognition}, vol. 07-12-June-2015, pp. 1--9, Jan. 2015.

\bibitem{Simonyan2015}
K.~Simonyan and A.~Zisserman, ``Very deep convolutional networks for
  large-scale image recognition,'' \emph{3rd International Conference on
  Learning Representations, ICLR 2015 - Conference Track Proceedings}, Jan.
  2015.

\bibitem{Szegedy2016}
C.~Szegedy, V.~Vanhoucke, S.~Ioffe, J.~Shlens, and Z.~Wojna, ``Rethinking the
  inception architecture for computer vision,'' in \emph{2016 IEEE Conference
  on Computer Vision and Pattern Recognition (CVPR)}.\hskip 1em plus 0.5em
  minus 0.4em\relax IEEE, Jun. 2016, pp. 2818--2826.

\bibitem{Mikolov2013}
T.~Mikolov, K.~Chen, G.~Corrado, and J.~Dean, ``Efficient estimation of word
  representations in vector space,'' \emph{1st International Conference on
  Learning Representations, ICLR 2013 - Workshop Track Proceedings}, Jan. 2013.

\bibitem{Smith2007}
R.~Smith, ``An overview of the tesseract ocr engine,'' in \emph{Proceedings of
  the International Conference on Document Analysis and Recognition, ICDAR},
  vol.~2, 2007, pp. 629--633.

\bibitem{Pennington2014}
J.~Pennington, R.~Socher, and C.~Manning, ``Glove: Global vectors for word
  representation,'' in \emph{Proceedings of the 2014 Conference on Empirical
  Methods in Natural Language Processing (EMNLP)}.\hskip 1em plus 0.5em minus
  0.4em\relax Stroudsburg, PA, USA: Association for Computational Linguistics,
  2014, pp. 1532--1543.

\bibitem{Peters2018}
M.~Peters, M.~Neumann, M.~Iyyer, M.~Gardner, C.~Clark, K.~Lee, and
  L.~Zettlemoyer, ``Deep contextualized word representations,'' in
  \emph{Proceedings of the 2018 Conference of the North American Chapter of the
  Association for Computational Linguistics: Human Language Technologies,
  Volume 1 (Long Papers)}.\hskip 1em plus 0.5em minus 0.4em\relax Stroudsburg,
  PA, USA: Association for Computational Linguistics, Jan. 2018, pp.
  2227--2237.

\bibitem{Eitel2015}
A.~Eitel, J.~T. Springenberg, L.~Spinello, M.~Riedmiller, and W.~Burgard,
  ``Multimodal deep learning for robust rgb-d object recognition,'' \emph{IEEE
  International Conference on Intelligent Robots and Systems}, vol. 2015-Decem,
  pp. 681--687, 2015.

\end{thebibliography}
